\documentclass[runningheads]{llncs}

\usepackage{eccv}

\usepackage{array}

\usepackage{eccvabbrv}

\usepackage{graphicx}
\usepackage{booktabs}
\usepackage[table]{xcolor}
\usepackage{amssymb}
\usepackage{threeparttable} 

\usepackage[accsupp]{axessibility}  
\usepackage{wrapfig}
\usepackage{subcaption}

\usepackage{hyperref}

\usepackage{orcidlink}
\usepackage{multirow}
\usepackage{multicol}

\newcommand{\ata}{\textcolor{red}{\bigstar}}
\newcommand{\atb}{\textcolor{blue}{\blacklozenge}}
\newcommand{\atc}{
\textcolor{green}{\bullet}
}
\newcommand{\atd}{
\textcolor{gray}{\blacksquare}
}
\newcommand{\ate}{
\textcolor{orange}{\blacktriangle}
}

\newcommand{\rg}[1]{\textbf{\textcolor{ForestGreen}{+#1\%}}}

\begin{document}

\title{Audio-Visual Camera Pose Estimation with Passive Scene Sounds and In-the-Wild Video} 

\titlerunning{Audio-Visual Camera Pose Estimation with Passive Scene Sounds}

\author{Daniel Adebi\orcidlink{0009-0000-1005-3374} \and
Sagnik Majumder\orcidlink{0000-0002-2851-1510} \and
Kristen Grauman\orcidlink{0000-0002-9591-5873}}

\authorrunning{D.~Adebi et al.}

\institute{The University of Texas at Austin, Austin TX 78712, USA \\ \email{\{ikadebi,sagnik,grauman\}@cs.utexas.edu}}

\maketitle

\begin{abstract}
 Understanding camera motion is a fundamental problem in embodied perception and 3D scene understanding. While visual methods have advanced rapidly, they often struggle under visually degraded conditions such as motion blur or occlusions. In this work, we show that passive scene sounds provide cues complementary to vision for relative camera pose estimation for in-the-wild videos. We introduce a simple but effective audio-visual framework that integrates direction-of-arrival (DOA) spectra and binauralized embeddings into a state-of-the-art vision-only pose estimation model. Our results on two large datasets show consistent gains over strong visual baselines, plus robustness when the visual information is corrupted. To our knowledge, this represents the first work to successfully leverage audio for relative camera pose estimation in real-world videos, and it establishes incidental, everyday audio as an unexpected but promising signal for a classic spatial challenge.
\end{abstract}

\section{Introduction}
\label{sec:intro}

Consider a video captured at a dark, crowded concert where someone wearing a camera moves through the venue searching for friends near the stage. The visual feed is severely degraded by dim lighting, motion blur, and constantly shifting crowds, making it difficult to track camera movement through vision alone. Yet the surrounding audio tells a clear spatial story: music grows louder as the person approaches the stage and the music shifts as they turn their head; conversations fade in and out with proximity; and ambient crowd noise provides persistent directional cues. We refer to such footage---video and audio captured under unconstrained, everyday conditions, without staging or simulation, and subject to natural visual and acoustic variation---as \emph{in-the-wild}.

Could a \emph{camera pose estimation model} use such incidental sounds in real-world video to refine its spatial story too?  This goal is appealing for multiple reasons.  First, incidental \emph{passive} scene sounds are compelling because---unlike actively emitted chirps for echolocation~\cite{yang2022camera,batvision,gao2020visualechoes}---they are non-intrusive, varied, and freely available with video.  Second, everyday sounds are invariant to some of the key obstacles for traditional vision-only camera pose estimation, like poor lighting, motion blur, textureless regions, and occlusions.  Audio's geometric cues remain reliable even when vision fails.

\begin{figure}[tb]
  \centering
  \includegraphics[width=0.78\linewidth]{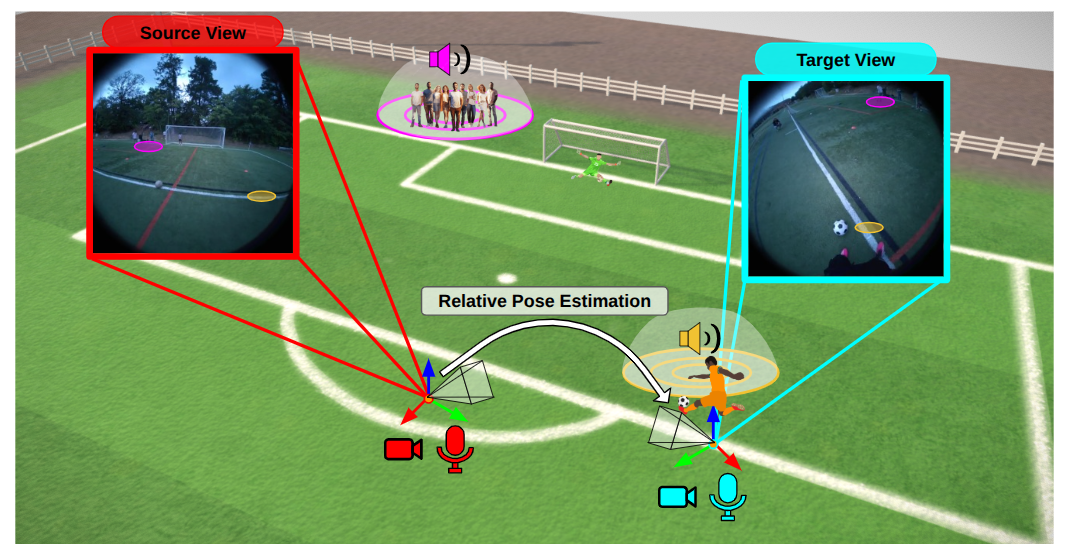}
  \caption{\textbf{Main idea}: we propose to estimate relative camera pose from in-the-wild videos using both vision and multichannel audio.  Unlike traditional active echolocation sensing, our model relies only on passive scene audio---gaining spatial cues opportunistically from naturally occurring ambient and foreground sound sources.
  }
  \vspace*{-0.1in}
  \label{fig:cvpr_teaser}
\end{figure}

Motivated by these observations, we introduce an audio-visual camera pose estimation model.  Given two video frames and their respective multi-channel (meaning at least two channels)  audio excerpts, the objective is to predict the relative camera pose separating the two frames. See \cref{fig:cvpr_teaser}.
Building on a state-of-the-art vision-only model~\cite{reloc3r}, we introduce a joint audio representation comprised of explicit direction-of-arrival audio cues and a mono-to-binaural ``lifted" embedding.  Importantly, we successfully train entirely with real-world video from a variety of scenarios and activities---a significant departure from the status quo, where methods exploring spatial audio rely on static (usually 3D-scanned~\cite{ramakrishnan2021hm3d,replica}) environments and simulated audio~\cite{slfm,yang2022camera,soundspaces2,gan,gao2020visualechoes,avfloorplan}.

We validate our approach on two datasets. Our model achieves significant improvements, outperforming the state-of-the-art~\cite{slfm} and establishes (to our knowledge) the first-ever results for audio-visual camera pose estimation on challenging real-world videos.  Our key contributions are:
\begin{itemize}
    \item We introduce a spatial audio encoder that learns spatial embeddings from two complementary cues: direction-of-arrival spectra and cross-view audio binauralization features.
    \item We validate our approach on the large-scale Ego-Exo4D dataset \cite{egoexo4d} containing hundreds of subjects and environments and the HM3D-SS \cite{ramakrishnan2021hm3d, soundspaces2} dataset, achieving state-of-the-art performance among audio-visual baselines for relative camera pose estimation.
    \item We provide a comprehensive analysis of how different audio characteristics affect multimodal camera pose estimation. 
\end{itemize}
Overall, our work establishes incidental, passive audio as a promising signal for camera pose.

\section{Related Work}

\vspace*{-0.1in}
\subsection{Vision-Based Camera Pose Estimation}
Accurate relative camera pose estimation is fundamental to a wide range of applications in 3D vision, augmented reality, and robotics. 
Most camera pose estimation methods rely solely on visual information~\cite{reloc3r, FAR, superglue, dust3r, monst3r, anycam, reposed, geonet, alligat0r, wang2025vggt, InterPose, wang2024eloftr, li2025megasam, cabon2025must3r, leroy2024grounding}, and limited prior work incorporates depth~\cite{monst3r, reposed}, IMU~\cite{Cerezo2024RGBDInertial}, or optical flow \cite{anycam}. 
Earlier methods such as SuperGlue \cite{superglue} and LoFTR \cite{loftr} rely on correspondence through keypoint matching, which makes them well suited to scenarios with high degrees of visual overlap between images. More recent approaches like DUSt3R \cite{dust3r} and Reloc3r \cite{reloc3r} leverage end-to-end transformer architectures, allowing direct correspondence estimation without relying on explicit keypoint detection and matching.
FAR \cite{FAR} integrates feature matching with learning-based approaches to improve performance under low visual overlap scenarios. 

All of these methods rely solely on visual input for camera pose prediction, leveraging
training data that spans simulated environments \cite{habitat19iccv, szot2021habitat, puig2023habitat3, InteriorNet18, replica}, reconstructed 3D scans of real scenes \cite{Matterport3D, dai2017scannet, scannet++, arkitscenes, blendedmvs}, photographs of static real-world environments \cite{phototourism, thomee2016yfcc100m, streetlearn, arnold2022mapfree, li2018megadepth, uco3d}, and real-world videos \cite{dl3dv, zhou2018stereo, xu2018youtube, epic-kitchens, waymo}. In our work, we incorporate audio as an additional modality to support pose estimation. 

\vspace*{-0.1in}
\subsection{Echolocation For Camera Poses}
To our knowledge, the only prior work to leverage audio for camera pose estimation does so in a very different setting---by actively emitting sounds into the environment in order to collect echoes~\cite{yang2022camera}---and is limited to training and testing in simulation.  Furthermore, no prior work uses passive scene audio for relative camera pose estimation.  Our method relies solely on naturally occurring sounds already present in the scene, without emitting signals, generating new sounds in a simulated scene, or requiring controlled acoustic conditions, making it passive, unobtrusive, and compatible with real-world videos.

\vspace*{-0.1in}
\subsection{Localizing Sound Sources}
Leveraging both sound and vision to localize sound sources has emerged as a powerful strategy~\cite{min2025supervising,wu2021binaural, Li_2024_CVPR,slfm}. 
Recent work shows how camera motion can reinforce estimates of sound source locations~\cite{min2025supervising,slfm}.  Results are done in simulation~\cite{slfm} or using real-world video egomotion to supervise binaural sound localization~\cite{min2025supervising}.
In related spatial problems, audio-visual models can learn floorplan maps~\cite{avfloorplan} and 3D scene structure~\cite{batvision} by associating sounding objects with likely rooms and/or listening for echolocation responses.  Other work trains navigation policies to move to a sounding object in an unmapped environment~\cite{gan,soundspaces2}. 
Unlike any of the above, our aim is to compute accurate 6 DoF relative camera poses, not sound source locations or environment maps.

\vspace*{-0.1in}
\subsection{Audio-visual Self-Supervised Pretraining}
Extensive prior work investigates how vision and audio interact to learn self-supervised representations, using the two modalities to create pretext tasks and improve training \cite{Ngiam2011MultimodalDL, Owens2016AmbientSP, Arandjelovi2017LookLA, korbar2018cooperative, owens2018audio, owens2016visually, afouras2020self, morgado2020learning}. These studies build tasks around synthesis \cite{Owens2016AmbientSP,owens2016visually}, cross‑modal alignment \cite{Arandjelovi2017LookLA, korbar2018cooperative, owens2018audio,afouras2020self,gao2020visualechoes}, and masked auto‑encoding approaches \cite{gong2023contrastive, georgescu2022audiovisual, huang2022mavil}, typically targeting semantic downstream problems such as audio‑visual event classification and retrieval. 
Different from these objectives, our approach incorporates spatial audio-based representations into camera pose estimation.

\section{
Task Motivation and Definition
}\label{sec:task}

The spatial sound perceived in a 3D scene is shaped by the relative location and orientation of the microphones used in capturing the sound, relative to the sound source(s), as well as the scene's geometry and materials. Consequently, any translation and/or rotation of the microphones results in changes in the sound's spatial attributes, which reveal the nature and extent of microphone motion. 
Based on this knowledge, we hypothesize that leveraging spatial audio \textbf{in addition to vision} for  relative camera pose prediction in in-the-wild videos---where the audio is captured using two or more microphones that move rigidly with the camera---can improve the pose prediction quality over using vision alone. We anticipate audio to be particularly valuable when vision alone is unreliable due to factors like occlusion, bad lighting, or camera failure, which are common in real-world video.  Meanwhile, a sound source that is not visible but is still audible adds useful cues---e.g., someone speaking or a radio located out of view.

\begin{figure*}[t]
  \centering
    \includegraphics[width=\linewidth]{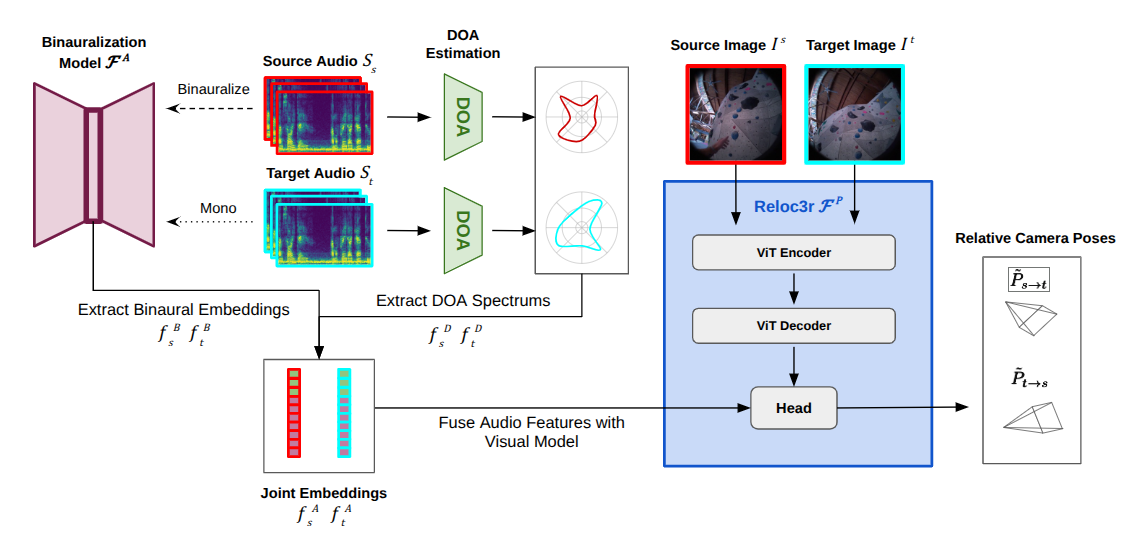}
  \vspace*{-0.25in}
  \caption{We extend the Reloc3r \cite{reloc3r} architecture with our Spatial Audio Encoder (SAE). Given a pair of source and target frames, coupled with their corresponding synchronized audio clips, our network predicts relative camera poses for the input image pair, in both directions (from source to target, and from target to source). The proposed SAE produces a combination of analytical and learned spatial audio embeddings that are subsequently integrated with Reloc3r's visual features 
  via late fusion, enabling the network to leverage complementary audio-visual cues for performing high-quality pose prediction. This design allows joint learning of translation and rotation and additionally provides robustness to visual corruption, as we show in results.}
\vspace*{-0.1in}
  \label{fig:full_method}
\end{figure*}

To validate our hypothesis, we propose a novel audio-visual in-the-wild relative camera pose estimation task.
In this task, we consider a video clip $V = (I, A)$, where $I$ and $A$ denote the visual and multi-channel audio streams, respectively. The visual stream $I$ consists of $N$ RGB image frames, such that $I = \big\{I_1, \ldots, I_N \big\}$. The audio stream $A$ is time-synchronous with $I$ and consists of $N$ audio segments, such that $A = \big\{A_1, \ldots, A_N\big\}$. Each audio segment $A_j$ has $C$ audio channels, such that $A_j = \big\{A_j^1, \ldots, A_j^C\big\}$, and is extracted by sampling a fixed-length time window (1000ms) centered at frame $I_j$, making $A_j$ temporally aligned with $V_j$. Note that the shorter the time window, the less likely there are substantial changes in  sound source positions intra-window. 

We make no assumptions about the audio other than it being captured from two (or more) microphones co-located with the camera.\footnote{We use the terms spatial audio and multi-channel audio interchangeably, to refer to audio captured from two or more microphones.} Such audio-visual capture is increasingly available in not only research devices like Aria~\cite{aria} glasses but also commercial consumer devices like GoPro, Insta360 Go, and RayBan Meta glasses. In particular, we are interested in leveraging \emph{passive} sounds---whatever are the incidental sounds in the scene from various sources, foreground or background---as opposed to \emph{active} sounds emitted into the environment to perform sensing, i.e., for echolocation.  The advantage of passive sounds is that they are non-intrusive and ``free", and available even on previously recorded video over which our models have no control.

Given a pair of source and target RGB frames $(I_s, I_t)$ and their corresponding multi-channel audio segments $(A_s, A_t)$, the goal in this task is to learn a model $\mathcal{F}$, such that $\mathcal{F}(I_s, A_s, I_t, A_t) = \tilde{P}_{s \rightarrow t}$, where $\tilde{P}_{s \rightarrow t}$ is an estimate of the 6-DoF relative camera pose $P_{s \rightarrow t}$ between $I_s$ and $I_t$. The matrix $P \in \mathbb{R}^{3 \times 4}$ represents the rotation and translation of the pose change. Formulating relative camera pose between two views is standard~\cite{reloc3r,  FAR, yang2022camera, slfm}. It also serves as a core primitive underlying recent multi-view and temporally-extended systems~\cite{dust3r,cabon2025must3r,wang2025vggt}, and is natural for online scenarios where 
future multi-view context is unavailable.

\vspace*{-0.1in}
\section{Approach}
\label{sec:method}

Our method is composed of two modules: \textbf{1)} a spatial audio encoder $\mathcal{F}^{{A}}$, and \textbf{2)} an audio-visual relative camera pose predictor $\mathcal{F}^{{P}}$. See \cref{fig:full_method}. While the audio encoder $\mathcal{F}^{{A}}$ is responsible for extracting rich spatial cues from the multi-channel audio inputs by using both analytical and learned representations, 
the pose predictor $\mathcal{F}^{{P}}$ combines these spatial audio cues with visual cues using a SOTA camera pose prediction backbone~\cite{reloc3r} to estimate the relative camera pose. Next, we describe the design of these modules and their training.

\subsection{Spatial audio encoder}\label{sec:audio_enc}

Given any multi-channel audio segment $A_j$ (cf. \cref{sec:task}), our spatial audio encoder $\mathcal{F}^{{A}}$ first computes a short-time Fourier transform (STFT) and represents the audio as a spectrogram $S_j$. $S_j$ is a matrix, such that $S_j \in \mathbb{R}^{C \times F \times W}$, where $F$ and $W$ are the number of frequency bins and time windows, respectively, and $C$ is the number of audio channels (e.g., $C=2$ for stereo or binaural audio). 

We compute two representations from the spectrogram: the direction-of-arrival (DOA) and a learned spatial audio embedding. For the former, we use an analytical direction-of-arrival (DOA) estimator~\cite{normmusic} to compute $f^D_j = \mathcal{F}^D (S_j)$, such that $f^D_j \in \mathbb{R}^{360}$. This DOA spectrum explicitly captures the distribution of sound energy around the microphone location, revealing the direction(s) of major sound sources in the environment with respect to the camera.  For the latter, we compute a binaural audio embedding  $f^B_j = \mathcal{F}^B(S_j)$ using a learned extractor (defined below) to provide an implicit but strong representation of the audio's spatial characteristics.

Finally, the encoder fuses $f^D_j$ and $f^B_j$ into a spatial audio embedding $f^A_j$.  The joint embeddings for the audio from both the source and target poses are passed  to the camera pose predictor $\mathcal{F}^P$ for further processing. Whereas the DOA feature $f^D_j$
signals the direction of major sound sources in the surroundings, 
the binaural audio features capture the full fabric of all overlapping sounds and their spatial layout. Importantly, even overlapping sound sources and ambient noises are a \emph{signal} for our camera pose estimation problem---not a nuisance as in traditional tasks like audio source separation. 
The synergy of the two complementary features, enabled by their fusion, is essential for high-quality pose prediction, as we show in results.  
Next, we elaborate on the design of these encoders and our feature fusion strategy. 

\medskip
\noindent \textbf{Direction-of-arrival spectrum.} For our direction-of-arrival (DOA) feature, we use the MUSIC~\cite{music} algorithm with frequency normalization~\cite{normmusic}, henceforth referred to as MUSIC++.  This method has been shown to produce  reliable DOA estimates in the presence of multiple sound sources. Given a multi-channel audio-spectrogram $S_j$, MUSIC++ analytically 
computes a one-dimensional DOA feature $f^D_j$, such that $f^D_j \in [0, 1]^{360}$ and each entry measures the average audio intensity across the corresponding $1^\circ$-wide sector in the azimuth circle. In essence, $f^D$ captures the angular variation in the audio intensity relative to the direction in which the microphones are facing. 
Notably, this DOA distribution will shift predictably as a function of the camera's motion---and vice versa---provided there are not sudden changes in the scene sounds between the time frame $I_s$ and $I_t$ are captured (1 sec.~in our experiments).

\medskip
\noindent \textbf{Learned binaural audio features.} Inspired by prior work for self-supervised audio(-visual)  learning~\cite{chen2023novel,vissound2.5,morgado,left-right,slfm,gao2020visualechoes,monotobinaural,Majumder_2024_CVPR}, we train an embedding to lift  monaural to binaural sound in the target viewpoint. The idea is to surface essential geometric cues from the audio. Specifically, we train $\mathcal{F}^B$ by attempting to solve a novel viewpoint\footnote{Here, ``viewpoint" refers to a microphone pose.} (view) acoustic synthesis (NVAS) task, where given a pair of views---one source and one target---the goal is to learn a model that can take the binaural audio for the source view and the monaural audio for the target view as inputs, and generate the binaural audio for the target view as output. By training the NVAS model on a large and challenging dataset~\cite{egoexo4d} comprising a large variety of dynamic 3D scenes with multiple overlapping sound sources at diverse locations and ambient noise, as well as arbitrary viewpoint changes, $\mathcal{F}^B$ learns to encode rich spatial audio features from two-channel audio.

\medskip
\noindent \textbf{Spatial audio feature fusion.} To leverage the complementary benefits offered by these two features, we fuse $f^D_j$ and $f^B_j$ into a single spatial audio feature $f^A_j$ through concatenation, such that $f^A_j = [f^D_j || f^B_j]$. Our fusion strategy, coupled with the use of non-linearities in the subsequent model layers (see \cref{sec:pose_predictor}), allows for full feature mixing, making the fused features highly expressive and conducive to high-quality camera pose estimation.

\subsection{Audio-visual relative camera pose predictor}\label{sec:pose_predictor}

For predicting the relative camera pose estimate $\tilde{P}$, we adapt a state-of-the-art vision-only relative camera pose predictor, Reloc3r~\cite{reloc3r}, such that it can take both visual and spatial audio inputs. Given a pair of RGB images $(I_s, I_t)$, the original Reloc3r model processes them using a pair of visual branches with the exact same architecture and shared weights, in order to predict their relative camera poses in both directions, $\tilde{P}_{s \rightarrow t}$ and $\tilde{P}_{t \rightarrow s}$. Towards that goal, it feeds each image to just one branch, which extracts visual features from the image using an encoder-decoder pair. Additionally, it uses cross-attention between the decoders of the two branches to establish fine-grained spatial correspondences between the corresponding visual features. It then uses a stack of linear layers as its camera pose prediction head, which leverages the aforementioned correspondences and estimates the camera pose of each input image relative to the other.  

To adapt Reloc3r for audio-visual inputs, we feed an audio-visual feature $f^{AV}$ to the camera pose estimation head of each branch, where $f^{AV}$ is produced by concatenating the head's original input feature $f^V$ and our spatial audio feature $f^A$, such that $f^{AV} = [f^A || f^V]$. Doing this late fusion of the visual and spatial audio features allows our pose predictor to learn important spatial correlations between the visual and audio modalities. Furthermore, it preserves the visual backbone’s cues: informative audio can augment vision, but uninformative
 audio has limited ability to override strong visual features.

\subsection{Model training}\label{sec:training}

\textbf{Binaural audio feature extractor training.} Following~\cite{vissound2.5, slfm}, given a target viewpoint $\mathcal{T}$ and its ground-truth binaural audio segment $A^B_{\mathcal{T}}$, our NVAS model, whose encoder we use for our binaural feature extractor $\mathcal{F}^B$, predicts an estimate $\tilde{S}^{\Delta}_{\mathcal{T}}$ of the spectrogram of the ground-truth difference between the left and right channels of the target audio,  $S^{\Delta}_{\mathcal{T}}$, such that  $S^{\Delta}_{\mathcal{T}}$ is computed by performing STFT on $A^l_{\mathcal{T}} - A^r_{\mathcal{T}}$. With this formulation, the model is not required to predict the exact value of each entry in spectrogram, which greatly speeds up training while still enabling the model to learn strong binaural features due to its requirement of accurately predicting any inter-channel differences. 
We train the NVAS model and consequently, our binaural feature extractor $F^B$, by computing $\mathcal{L}^B$, where $\mathcal{L}^B$ is the $L_1$ loss between $S^{\Delta}_{\mathcal{T}}$ and $\tilde{S}^{\Delta}_{\mathcal{T}}$, such that
\begin{align}
    \mathcal{L}^B = || \tilde{S}^{\Delta}_{\mathcal{T}} - S^{\Delta}_{\mathcal{T}} ||_1.
\end{align}

\noindent \textbf{Camera pose estimator training.} Given an RGB image pair $(I_s, I_t)$ and the corresponding relative camera pose ground-truths $(P_{s \rightarrow t}, P_{t \rightarrow s})$, we train our audio-visual camera pose predictor $F^P$ using $\mathcal{L}^P$, where $\mathcal{L}^P$ is the mean of the pose prediction loss $L^P_{s \rightarrow t}$ for $P_{t \rightarrow s}$ and its converse $L^P_{t \rightarrow t}$, corresponding to the prediction loss for $P_{t \rightarrow s}$, such that 
\begin{align}
    \mathcal{L}^P = L^{P}_{s \rightarrow t} + L^P_{t \rightarrow s}.
\end{align}
Following~\cite{reloc3r}, our pose prediction loss $L^P$ has two components, measuring the error in the rotation and translation predictions, respectively.

\section{Experiments}
\label{sec:experiments}
We present experiments validating our model on three datasets, (1) the in-the-wild video dataset Ego-Exo4D~\cite{egoexo4d}, (2) the perceptually realistic HM3D-SoundSpaces~\cite{soundspaces2, ramakrishnan2021hm3d}, which allows comparison against the SOTA audio-visual method, and (3) test sets with added visual corruptions for both.

\subsection{Experimental Setup}
\label{sec:experimental-setup}
\noindent \textbf{Datasets.} To assess our method’s performance on real-world data, we first train and evaluate our models on the \textbf{Ego-Exo4D dataset} \cite{egoexo4d}, using all 132 hours of egocentric sequences for which there are audio recordings (Ego-Exo4D's exo cameras are static and hence not relevant for pose estimation).  Audio in Ego-Exo4D is captured with 7 microphones on the Aria glasses, which provide synchronized multichannel audio and precise camera pose annotations. The data spans a diverse range of scenarios, motions, and activities, including indoor navigation, social interactions, and object manipulation, enabling evaluation across varied acoustic conditions like speech vs. object sounds, and foreground sounds vs. ambient noise, and visual conditions such as occlusions and lighting changes. We note that this complex video content pushes the boundary compared to how relative camera pose is typically assessed in simulated settings and/or with static scenes. See \cref{sec:additionalresults} in supplemental for more detailed analysis on the camera pose and audio distribution statistics.

To generate our dataset, we uniformly sample frame pairs that are one second apart, ensuring sufficient motion while maintaining temporal continuity. This sampling choice offers a consistent supervision signal without restricting generality, as the model is trained across a wide variety of activities, environments, and motion magnitudes. We also evaluate our model performance on different temporal splits in \cref{tab:temporal_splits} (Supp.) to showcase how the model generalizes to different time gaps between input frames.
We follow the official Ego-Exo4D data splits, ensuring that frame pairs used for training and testing are sampled from entirely disjoint video subsets, fully preventing cross-split overlap during evaluation.

Secondly, to enable direct comparison with SLfM~\cite{slfm}, which generates camera orientation and sound source locations, we additionally train and evaluate a variant of our approach on the Habitat-Matterport3D SoundSpaces \textbf{(HM3D-SS) dataset} \cite{ramakrishnan2021hm3d, slfm}, which combines Habitat-Matterport3D's photorealistic indoor scenes with the SoundSpaces \cite{soundspaces2} framework in order to produce realistic spatial audio in simulation. This variant uses 2 microphones instead of 7. We follow the same data splits and scene configurations used by SLfM~\cite{slfm}.

Finally, to assess robustness to corrupted visual inputs, we also generate a \textbf{corrupted test set} for each of the above datasets
by applying a random combination of photometric and geometric transformations to each frame, where both the chosen transformations and their strengths are randomly decided. The set of possible perturbations includes Gaussian blur, Gaussian noise, and color jitter (brightness, contrast, saturation, and hue adjustments). See \cref{tab:corrupt} for examples. These degradations include both standard corruption types used in prior robustness work \cite{hendrycks2019benchmarking, mintun2021on, liu2024benchmarking, Ma2024PoseBenchBT} as well as more extreme perturbations that occur when the transformations are applied with full strength to simulate unexpected visual failures, enabling evaluation under both typical and worst-case visual conditions.

\medskip
\noindent \textbf{Baselines.} We compare against the following baselines and state-of-the-art methods: 
\begin{itemize}
    \item \textbf{Chance:} this is a heuristic that simply outputs the mean of the ground-truth relative camera poses from the validation set 
    \item \textbf{Naive Audio Camera Pose (NA-CP):} this baseline predicts the relative camera pose by replacing the RGB inputs to a Reloc3r~\cite{reloc3r} model with corresponding multi-channel audio spectrograms, and finetunes it to predict estimates of the corresponding ground-truth camera poses.
    \item \textbf{Direction-of-arrival Camera Pose (DOA-CP):} this baseline first computes  the direction-of-arrival (DOA) spectrum, and then predicts the relative camera pose by processing the DOA spectrum using a stack of linear layers.
    \item \textbf{Reloc3r~\cite{reloc3r}:} a state-of-the-art vision-only relative camera pose predictor that takes pairs of RGB images as inputs and uses a transformer encoder-decoder model to predict the camera pose for the image pair, in both directions.
    \item \textbf{Reloc3r~\cite{reloc3r}-AV:} an audio-visual version of Reloc3r that performs late fusion (cf. \cref{sec:pose_predictor}) of Reloc3r's visual features and the audio features from the corresponding layer of our NA-CP baseline, and feeds the fused features to Reloc3r's pose prediction head to estimate the relative camera pose. We finetune this model using pairs of audio-visual inputs and relative camera pose ground-truths.
    \item \textbf{Reloc3r~\cite{reloc3r}-AV Cross Attention:} we leverage multimodal cross-attention to integrate audio and visual features, allowing the model to selectively fuse information across modalities for improved scene understanding. This approach is inspired by recent work in audio-visual learning, where cross-attention enables effective alignment between visual cues and audio signals \cite{venkatraman2024multimodal, lee20252st}.
    \item \textbf{Ours w/ Monaural:} an ablation of our approach where we modify our binauralization model so that it predicts single-channel audio. From here, we extract the monaural embeddings from our audio encoder to fuse with the rest of our model. DOA spectra are still used as an input to this model ablation.
\end{itemize}

\begin{table*}[!tb]
\centering
\tiny
\caption{Comparison of relative camera pose estimation performance on Ego-Exo4D dataset validation split. We report AUC@5, AUC@10, and AUC@20 metrics for rotation only, translation only, and total based on the maximum of the rotation and translation error. Best performing results are in bold, and second best results are underlined. Relative gains are based on performance over vision-only model. All gains over baselines and ablations are statistically significant ($p \leq$ 0.05). }
\setlength{\tabcolsep}{2pt}
\begin{tabular}{l|ccc|ccc|ccc}
\toprule
\multirow{2}{*}{Methods} & \multicolumn{3}{c|}{Rotation Only} & \multicolumn{3}{c|}{Translation Only} & \multicolumn{3}{c}{Total} \\
 & AUC@5 & AUC@10 & AUC@20 & AUC@5 & AUC@10 & AUC@20 & AUC@5 & AUC@10 & AUC@20 \\
\midrule
\multicolumn{10}{l}{\textit{Vision Only}} \\
\midrule
Reloc3r~\cite{reloc3r} & 37.35  & 53.62  & 68.98  & 0.73 & 2.63 & 7.77 & 0.57  & 2.33 & 7.33  \\
\midrule
\multicolumn{10}{l}{\textit{Audio Only }} \\
\midrule
Chance & \underline{19.02}  & \underline{33.33} & 50.03 & 0.06 & 0.26  & 1.11 & 0.02 & 0.12 & \underline{0.76} \\
NA-CP Model & 18.83 & 33.27 & \underline{50.05} & \underline{0.07} & \underline{0.29} & \underline{1.16} & \underline{0.02} & \underline{0.12} & 0.72  \\
DOA CP Model & \textbf{19.07} & \textbf{33.37} &\textbf{50.07} & \textbf{0.09} & \textbf{0.37} & \textbf{1.38} & \textbf{0.03} & \textbf{0.14} & \textbf{0.78} \\
\midrule
\multicolumn{10}{l}{\textit{Audio + Vision}} \\
\midrule
Reloc3r~\cite{reloc3r}-AV & 37.10 & 53.30 & 68.72 & 0.88 & 2.97 & 8.53 & 0.69 & 2.65 & 8.07 \\ 
Reloc3r~\cite{reloc3r}-AV Cross-Att & \underline{38.01} & 53.82 & 69.01 & 0.92 & 3.15 & 8.89 & \underline{0.76} & 2.83 & 8.34 \\
Ours w/ Monaural & 37.98 & 53.80 & 68.98 & 0.94 & 3.18 & 8.89 & 0.74 & 2.82 & 8.37 \\
Ours w/o DOA  & 37.48 & 53.45 & 68.88 & \underline{0.96} & \underline{3.21} & \underline{8.92} & 0.75 & \underline{2.86} & \underline{8.41}  \\ 
Ours w/o Binaural & 37.99 & \underline{53.87} & \underline{69.04} & 0.93 & 3.12 & 8.80 & 0.74 & 2.80 & 8.34 \\
Ours &\textbf{38.54} & \textbf{54.49} & \textbf{69.61} & \textbf{1.03} & \textbf{3.36} & \textbf{9.34} &\textbf{0.81} & \textbf{2.99} & \textbf{8.82}  \\
\textit{Relative Gains (\%)} & \rg{3.2} & \rg{1.6} & \rg{0.9} & \rg{41.1} & \rg{27.8} & \rg{20.2} & \rg{42.1} & \rg{28.3} & \rg{20.3} \\
\midrule
\multicolumn{10}{l}{\textit{Corrupted Vision}} \\
\midrule
Reloc3r~\cite{reloc3r}  & 25.17 & 38.88 & 52.80 & 0.34 & 1.28 & 4.28 & 0.21 & 1.00 & 3.66 \\
\midrule
\multicolumn{10}{l}{\textit{Audio + Corrupted Vision}} \\
\midrule
Reloc3r~\cite{reloc3r}-AV  & 26.17 & 39.76 & 53.59 & 0.43 & 1.57 & 4.96 & 0.29 & 1.27 & 4.36 \\
Reloc3r~\cite{reloc3r}-AV Cross-Att & 26.38 & 39.83 & 53.68 & 0.42 & 1.61 & 5.08 & \underline{0.32} & 1.29 & 4.44 \\
Ours w/ Monaural & \underline{27.00} & \underline{40.28} & \underline{53.89} & 0.44 & 1.61  & 5.07  & 0.31 & 1.28  & 4.42 \\
Ours w/o DOA & 26.14 & 39.76 & 53.71 & \underline{0.45} & 1.61 & 5.06 & 0.31 & 1.29 & 4.42 \\
Ours w/o Binaural & 26.56 & 40.06 & 53.81 & 0.44 & \underline{1.62} & \underline{5.10} & 0.29 & \underline{1.30} & \underline{4.48} \\
Ours & \textbf{27.17} & \textbf{40.66} & \textbf{54.46} & \textbf{0.49} & \textbf{1.77} & \textbf{5.45} & \textbf{0.35} & \textbf{1.44} & \textbf{4.80} \\
\textit{Relative Gains (\%)} & \rg{8.0} & \rg{4.4} & \rg{3.1} & \rg{44.1} & \rg{38.3} & \rg{27.3} & \rg{66.7} & \rg{44.0} & \rg{31.2} \\
\bottomrule
\end{tabular}
\label{tab:egoexo4d_results}
\end{table*}

\noindent \textbf{Evaluation metrics.} Following previous research \cite{reloc3r, superglue, loftr, wang2024eloftr}, we evaluate our model on the Ego-Exo4D dataset using 
AUC@5/10/20. These metrics compute the area under the accuracy curve (AUC) for camera pose estimation, with thresholds of 5, 10 and 20 degrees on the maximum angular error in the rotation and translation predictions.
This provides a comprehensive measure of relative pose accuracy across varying error tolerance levels. For the HM3D-SS dataset, we strictly follow the evaluation protocol used in prior work~\cite{slfm} and  report the mean absolute error (MAE) in predictions of the rotation azimuth.

\begin{table}[t]
\tiny
  \caption{Results on the HM3D-SS dataset test split. We report rotation accuracy using mean absolute error (MAE ($^\circ$)) for the original data (left)  and corrupted visual signals (right), following the evaluation protocol of SLfM \cite{slfm}. Lower values are better. Relative performance gains are based on performance over vision-only model. All gains over baselines and ablations are statistically significant ($p \leq$ 0.05).
  }
  \label{tab:hm3d-ss}
  \centering
  \begin{tabular}{@{}l|c|c@{}}
    \toprule
   Method & Original & Corrupted \\
    \midrule
   Chance & \multicolumn{2}{c}{29.41}  \\
   SLfM \cite{slfm} & 0.77 & --- \\
     \midrule
   NA-CP Model & \multicolumn{2}{c}{2.56}  \\
   DOA-CP Model & \multicolumn{2}{c}{28.63} \\
   \midrule
   Reloc3r~\cite{reloc3r} & 0.09 & 3.83 \\
   Reloc3r~\cite{reloc3r}-AV  & 0.09 & 3.28 \\
   Reloc3r~\cite{reloc3r}-AV Cross-Att & 0.08 & 2.82 \\
   Ours w/ Monaural & 0.08 & 2.89 \\
   Ours w/o DOA  & 0.08 & 2.78\\ 
   Ours w/o Binaural & 0.08 & 2.66\\
   Ours & \textbf{0.06} & \textbf{2.46} \\
   \textit{Relative Gains (\%)} & \rg{33.3} & \rg{35.8} \\
    \bottomrule
  \end{tabular}
\end{table}

\subsection{Results}
\label{sec:results}
We present the results for both datasets, including their visually corrupted variants, and then provide analysis of what audio properties are most and least amenable to augmenting vision for camera pose estimation.

\medskip
\noindent \textbf{Audio-visual relative camera pose estimation.} \Cref{tab:egoexo4d_results} reports relative camera pose estimation results on the Ego-Exo4D test split.\footnote{Note that the SLfM~\cite{slfm} method is not applicable to Ego-Exo4D because when it jointly trains for sound source localization and camera motion, it requires ground truth direction-of-arrival values, which Ego-Exo4D does not provide.}
Among audio-only models, the DOA-CP model achieves the strongest overall performance, with higher AUC scores across most thresholds compared to both the mean baseline and the naive audio model.  
This demonstrates that direction-of-arrival (DOA) features extracted from multichannel raw audio provide meaningful geometric cues for estimating relative pose, even without visual supervision.

When combining audio information with visual inputs, our approach outperforms all of our baselines and ablations with statistical significance ($p \leq$ 0.05), with significant relative gains in total AUC over all thresholds (42\%/28\%,/20\% for AUC@5/10/20 respectively). We can see that incorporating structured audio representations leads to consistent improvements across both rotation and translation metrics. Specifically, the addition of DOA features substantially enhances rotation accuracy, reflecting their ability to encode spatial geometry complementary to vision, while the binaural embeddings provide cues that improve translation estimates. Our approach using monaural embeddings worsens performance, showcasing the importance of inferred binauralization in our method; the graceful degradation highlights its practical utility for even single-microphone devices. Our full model achieves the highest total AUC across all thresholds, demonstrating the benefits of jointly leveraging both embeddings.

\begin{table*}[t]
\centering
\caption{AUC@20 scores demonstrating robustness under different visual corruption scenarios, with corresponding visualizations below. Our method consistently outperforms the baseline for all forms of corruption with statistical significance ($p < 0.05$).}
\tiny
\setlength{\tabcolsep}{2pt}
\begin{minipage}{0.32\linewidth}
\centering
\begin{tabular}{c|ccc}
\hline
\textbf{$\sigma$ Noise} & \textbf{Vision Only} & \textbf{Ours} \\
\hline
0.00 & 7.33 & \textbf{8.82} \\
0.05 & 6.29 & \textbf{7.62} \\
0.10 & 4.82 & \textbf{5.92} \\
0.20 & 4.28 & \textbf{5.55} \\
0.30 & 4.08 & \textbf{5.52} \\
\hline
\end{tabular}

\vspace{2mm}
\includegraphics[width=0.9\linewidth]{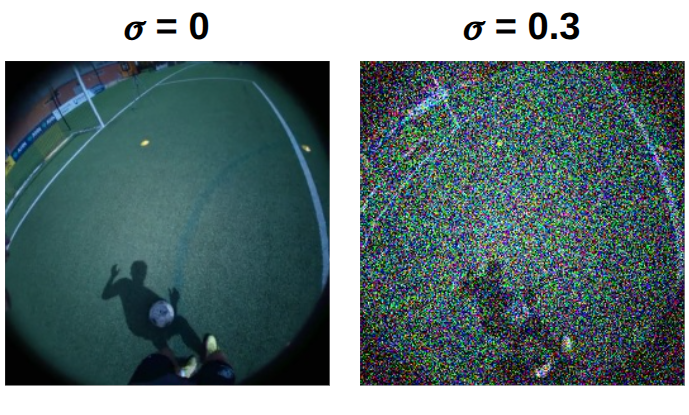}  
\caption*{(a) Gaussian Noise}
\end{minipage}
\hfill
\begin{minipage}{0.32\linewidth}
\centering
\begin{tabular}{c|ccc}
\hline
\textbf{$\sigma$ Blur} & \textbf{Vision Only} & \textbf{Ours} \\
\hline
0 & 7.33 & \textbf{8.82} \\
1 & 5.82 & \textbf{7.61} \\
2 & 5.07 & \textbf{6.21} \\
4 & 3.52 & \textbf{4.41} \\ 
8 & 2.63 & \textbf{3.47} \\
\hline
\end{tabular}

\vspace{2mm}
\includegraphics[width=0.9\linewidth]{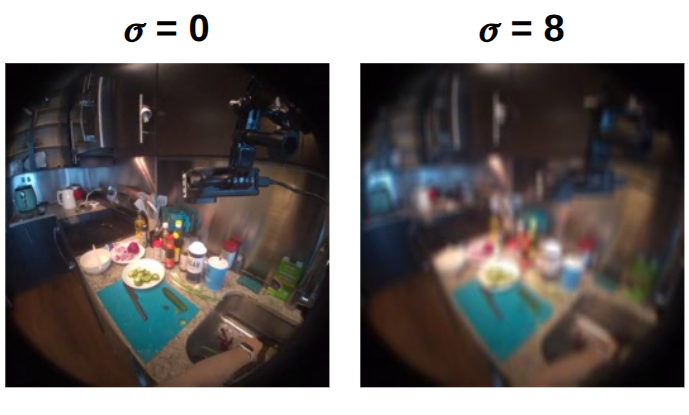}
\caption*{(b) Gaussian Blur}
\end{minipage}
\hfill
\begin{minipage}{0.32\linewidth}
\centering
\begin{tabular}{c|cc}
\hline
\textbf{Jit. Strength} & \textbf{Vision Only} & \textbf{Ours} \\
\hline
0.0 & 7.33 & \textbf{8.82} \\
0.2 & 7.27 & \textbf{8.79} \\
0.4 & 6.55 & \textbf{7.97} \\
0.6 & 6.32 & \textbf{7.67} \\
0.8 & 6.20  & \textbf{7.64} \\
\hline
\end{tabular}

\vspace{2mm}
\includegraphics[width=0.9\linewidth]{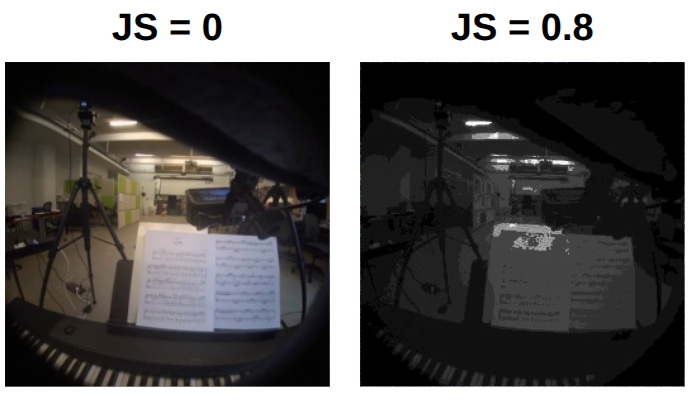}
\caption*{(c) Color Jitter}
\end{minipage}
\label{tab:corrupt}
 \vspace*{-0.35in}
\end{table*}

Furthermore, \cref{tab:egoexo4d_results} shows that under corrupted vision conditions (bottom third of the table), our model maintains significantly higher performance than all baselines, highlighting its robustness when visual cues are degraded. These results collectively confirm that our multi-embedding fusion not only improves overall accuracy but also enhances cross-modal reliability in challenging visual environments.  
\Cref{tab:corrupt} breaks down the relative accuracy for our method and its vision-only counterpart for multiple levels of different categories of corruption. Note that we consistently achieve higher scores than the vision-only baseline across all forms of degradation at all corruption levels.

Having analyzed the impact of visual corruption, we next examine the impact of audio corruption at inference time. Unlike in source separation, our method does not depend on clean or intelligible audio content; rather, it gains signal from any inter-channel differences, so additional sound energy, even if perceptually messy, still offers usable spatial cues. Indeed, we find that with noisy audio (Gaussian noise), performance remains largely preserved (Total AUC@5/10/20 = 0.78/2.88/8.62), well above the vision-only baseline (0.57/2.33/7.33; \cref{tab:egoexo4d_results}). This is again consistent with our late-fusion design (\cref{sec:pose_predictor}), which limits the influence any single corrupted modality can have on the final prediction.

\begin{figure*}[!tb]
\centering
\includegraphics[width=\textwidth]{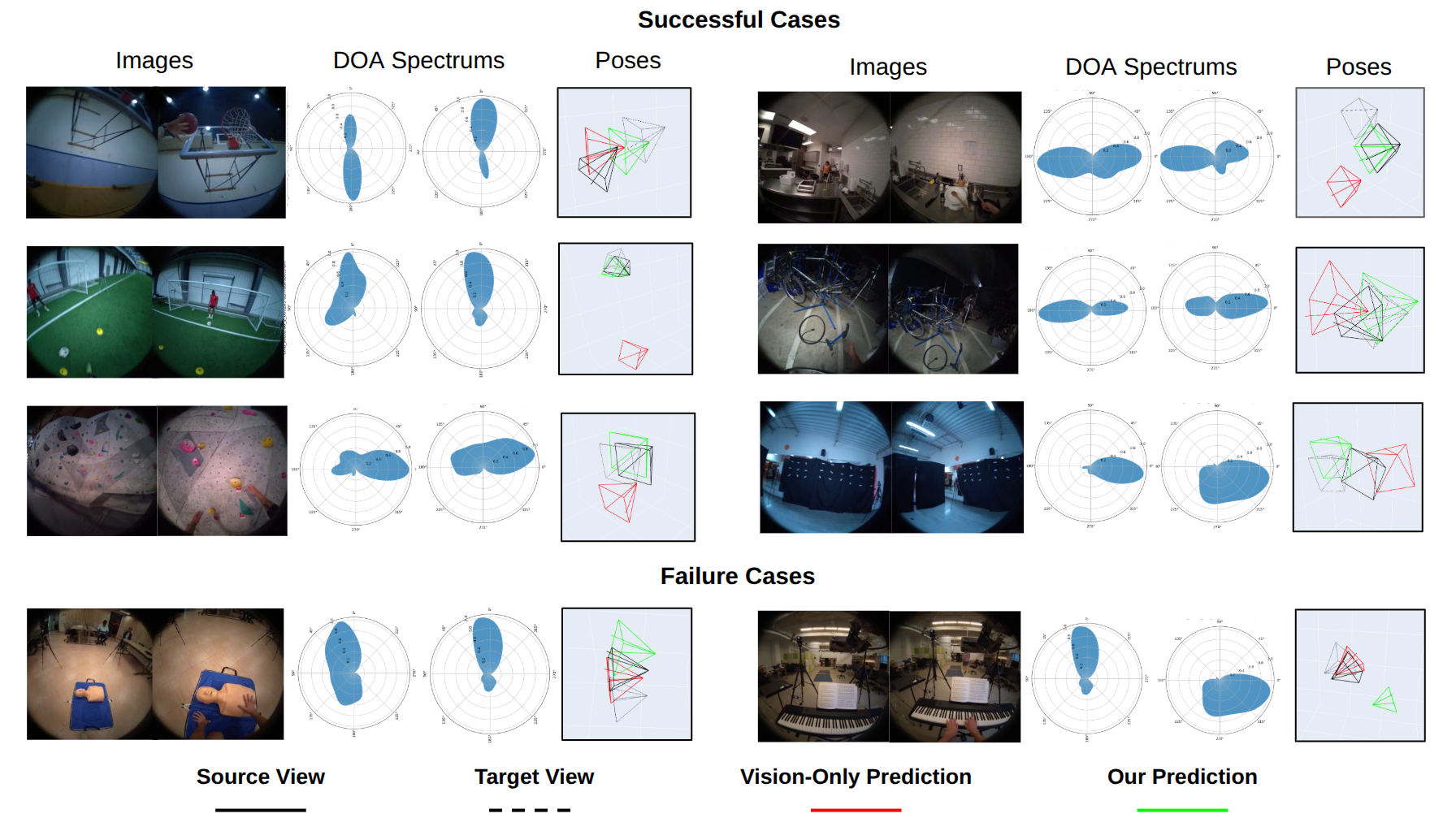}
\caption{Qualitative examples of our audio-visual method outperforming a vision-only state-of-the-art relative camera pose estimation model. Our model performs best when there are consistent audio signals in the scene that aid in localization, such as the soccer video in the second row. Some failure cases of our model include when a single dominant audio starts up in one frame but does not exist in a previous frame (e.g., bottom right, where the musician doesn't start playing music til the second frame).}
\label{fig:qual}
\end{figure*}

\Cref{fig:qual} provides some qualitative examples comparing our model to the vision-only baseline. Here we see the impact that ambient audio has across many scenarios. Our method can adapt to sudden changes in pose that seem to throw off our vision-only model. There are a few cases where our method struggles, though, specifically when a single dominant audio starts up (or stops happening) between frames.

\Cref{tab:hm3d-ss} shows the results on the HM3D-SS dataset, for both the original data (left) and the visually corrupted (right).  We can see our model achieves the lowest mean absolute rotation error, outperforming both  the state-of-the-art SLfM~\cite{slfm} and the Reloc3r~\cite{reloc3r} baseline in both normal and corrupted visual settings. While the improvement over the visual-only baseline is minimal, this is largely due to the controlled nature of HM3D–SS scenes, where the visual signal alone is highly informative.  This reinforces our emphasis on real-world video as a more challenging and essential testbed.

We also observe that the DOA Camera Pose model performs poorly in this setting, which may stem from inaccuracies in how direction-of-arrival features are computed in simulation and the absence of precise microphone position parameters in the SoundSpaces~\cite{soundspaces2} configuration. Nonetheless, our DOA-enhanced fusion model consistently surpasses the naive audio variant and vision-only baseline, suggesting that DOA cues remain beneficial even under imperfect calibration and strong visual dominance.

\begin{table*}[t]
\centering
\tiny
\caption{Analysis of multimodal performance under controlled motion/audio regimes (left) and across diverse Ego-Exo4D scenarios (right).}
\label{tab:combined}
\begin{subtable}[t]{0.48\textwidth}
\centering
\caption{AUC@5/10/20 under varying camera motion and acoustic change settings.}
\label{tab:decoupled}
\begin{tabular}{l|ccc}
\toprule
Method & AUC@5 & AUC@10 & AUC@20 \\
\midrule
\multicolumn{4}{c}{\textit{Large Translation >$1$m, Small Rotation <$10^\circ$}}\\
\midrule 
Vision-Only & 2.56  & 7.58 & 16.65 \\
Ours & \textbf{3.57} & \textbf{9.51} & \textbf{20.08} \\
\midrule
\multicolumn{4}{c}{\textit{Small Camera Motion, Large Sound Change}} \\
\midrule
Vision-Only & 0.71 & 2.95 & 9.15 \\
Ours & \textbf{1.05} & \textbf{3.74} & \textbf{10.89} \\ 
\midrule
\multicolumn{4}{c}{\textit{Large Camera Motion, Small Sound Change}} \\
\midrule
Vision-Only & 0.30 & 1.36 & 4.99 \\
Ours & \textbf{0.40} & \textbf{1.57} & \textbf{5.89} \\ 
\bottomrule
\end{tabular}
\end{subtable}
\hfill
\begin{subtable}[t]{0.48\textwidth}
\centering
\caption{Impact of scene audio characteristics across scenarios.  The middle column lists each scenario's predominant audio characteristic(s), identified via manual inspection. The rightmost column reports the percentage of frames where our method outperforms the vision-only (Reloc3r) baseline.}
\label{tab:ambient-audio}
\begin{tabular}{|c|c|c|}
\hline
\textbf{Scenario} & \textbf{Audio type(s)} & \textbf{Audio helps? (\%)} \\
\hline
Soccer & $\ata$ $\atd$ $\ate$ & 84.63\\
Music & $\atb$ $\atc$ & 81.40 \\
Bike Repair & $\atb$ $\atd$ $\ate$ & 79.55  \\ 
Rock Climbing & $\ata$ $\atb$ $\atc$ & 79.41\\
Basketball & $\ata$ $\atb$ $\ate$ & 72.15\\
Health & $\atb$ $\atd$ & 67.16 \\
Dance & $\atc$ $\ate$ & 66.14 \\ 
Cooking & $\atb$ $\atd$ & 64.13\\
\hline
\multicolumn{3}{|p{5.5cm}|}{
\tiny
\textbf{Audio Types:}
$\textcolor{red}{\bigstar}$ Far-field \,
$\textcolor{blue}{\blacklozenge}$ Near-field \,
$\textcolor{green}{\bullet}$ Dominant Single \,
$\textcolor{gray}{\blacksquare}$ Low Signal \,
$\textcolor{orange}{\blacktriangle}$ Frequent Changes
} \\
\hline
\end{tabular}
\end{subtable}
\vspace*{-0.2in}
\end{table*}
\medskip
\noindent \textbf{Decoupling camera motion and audio change.}
\Cref{tab:decoupled} evaluates performance in regimes where camera motion and acoustic change are explicitly decoupled, and in heavy translation settings. To construct these subsets, we automatically filter sequences using camera pose statistics, defining large motion as translation over 1m or rotation greater than $20^\circ$ (above the mean of the dataset). We separately quantified sound variation to detect large acoustic changes (e.g. a sound exists in one frame but does not exist in second frame). We then cross-selected clips satisfying one condition while constraining the other, and performed a light manual verification pass to ensure clean separation between motion- and audio-dominant cases.  

Across these controlled settings, our method consistently outperforms Vision-Only. Gains are particularly evident when sound sources change substantially, even under large camera motion, indicating robustness to acoustic variability. Improvements also persist when camera motion exceeds 1m translation but remains under 10° rotation, demonstrating strong performance in high-translation, low-rotation regimes where directional audio cues provide especially informative constraints on relative displacement.

\medskip
\noindent \textbf{Analysis of passive in-the-wild scene audio characteristics.} \Cref{tab:ambient-audio} summarizes representative audio characteristics observed across the Ego-Exo4D scenarios and illustrates that the proposed approach consistently improves performance across multiple real-world acoustic settings represented in the dataset. For each scenario, we annotate the predominant audio characteristics most frequently observed in the data, including far-field source audio (distant sound sources), near-field or egocentric audio (sounds proximate to the camera), dominant single sound sources, low or minimal audio signals, and frequent changes in sound content between consecutive frames. We then measure the frequency at which our method outperforms the vision-only state-of-the-art baseline (``Audio Helps? (\%)'' column). For more details on how we annotate our audio characteristics, see \cref{tab:audio-distribution} in the supplemental.

Our gains are strongest in scenes with multiple spatially distributed sound sources, where DOA spectra and binauralized embeddings provide complementary geometric cues to vision. Scenarios with the most concurrent sources yield our most frequent gains (80-85\%, \cref{tab:audio-distribution} in Supp). Even under omnidirectional sound (flat DOA spectra), our binaural embeddings still allow gains over vision only by 8\%/6\%/3\% for AUC@5/10/20. In Soccer and Music, concurrent sources (e.g., ball impacts, footsteps, instruments, people talking) generate distinct directional signatures that evolve with camera motion, enabling our audio-visual model to better resolve relative pose—especially under visual degradation. Bike Repair similarly benefits from spatially separated tool interactions. In contrast, Cooking and Dance often contain fewer independently localized emitters, more centralized acoustic activity, or a single dominant sound source, reducing directional diversity, though consistent sound cues still yield meaningful improvements over vision-only baselines. Further analysis can be found in \cref{sec:additionalresults} in the supplemental.

\medskip

\noindent
\textbf{Does audio truly provide a geometric signal?} To investigate whether the audio branch exploits geometrically meaningful cues rather than scene-, activity-, or dataset-specific correlations, we perform a series of negative-control experiments that progressively disrupt the correspondence between audio and camera motion. Performance degrades as the audio becomes increasingly inconsistent with the underlying scene geometry: synchronized audio achieves the highest AUC@5 (0.81), followed by temporally offset audio (1–5s; 0.68), same-scene shuffled audio (0.66), and cross-scene audio replacement (0.50), which performs worse than the vision-only baseline (0.57). These results indicate that dataset priors and semantic correlations are not enough---
our results depend critically on the geometric consistency between the audio observations and the visual scene. 

\medskip

\noindent
\textbf{Failure modes.} While our audio-visual framework demonstrates strong robustness across diverse scenarios, it remains susceptible to specific failure modes, particularly when acoustic cues become disjointed from the underlying scene geometry. Our model can struggle in near-silent or diffuse-source conditions where directional cues are inherently weak. Furthermore, because the framework relies on the geometric consistency of the audio, artificially disrupting this alignment, e.g. inserting an unrelated soundtrack in post-production, would actively mislead the model and degrade performance. Abrupt acoustic changes that occur independently of camera motion also present a challenge (see \cref{fig:qual}, bottom row).

\section{Conclusion}
We present a multimodal approach for relative camera pose estimation in video that exploits spatial cues from multichannel audio. Our method achieves state-of-the-art performance among audio-visual baselines on both real-world and simulated settings, demonstrating that ambient audio contains rich spatial structure to complement visual signals for a classic spatial vision task.

While our approach is effective, it has some limitations that present opportunities for future work. Translation estimation remains challenging in visually degraded or acoustically complex scenarios. In addition, it would be interesting to extend the model to explicitly gate the audio's influence depending on its properties (e.g., stationarity). Exploring cross-dataset and cross-device generalization is also an important area for future work. Despite these challenges, our work establishes that audio-visual fusion substantially improves pose estimation across diverse acoustic conditions. We hope this work motivates further research in leveraging sound for geometric reasoning and robust scene understanding.

\section*{Acknowledgements}

This research is supported in part by  Lyda Hill Philanthropies.
We thank the UT Austin Vision Group, Ziyang Chen (U. Michigan), and Kazuki Shimada (Sony) for helpful discussions.

\bibliographystyle{splncs04}
\bibliography{main}

\clearpage
\begin{center}
{\Large \bfseries Supplementary Material}
\end{center}
\setcounter{page}{1}

\vspace{0.5em}
\noindent\textbf{Supplementary Table of Contents}
\begin{itemize}
    \item[] \textbf{\ref{sec:additionalresults}} More Analyses
        \begin{itemize}
            \item[] \textbf{\ref{sec:additionalmodelanalyses}} Additional Model Analyses
            \item[] \textbf{\ref{sec:furtheregoexo}} Further Ego-Exo4D Dataset Analysis
        \end{itemize}
    \item[] \textbf{\ref{sec:impdetails}} Implementation Details
    \item[] \textbf{\ref{sec:hardware}} Hardware and Resource Constraints
    \item[] \textbf{\ref{sec:ethics}} Ethical Considerations
\end{itemize}

\begin{table}[!ht]
    \caption{Comparison of pose estimation performance (Total AUC scores) when inertial measurements are available on Ego-Exo4D dataset \cite{egoexo4d}. While IMUs provide strong motion cues, incorporating spatial audio further improves accuracy, indicating that passive scene sounds provide complementary information beyond visual and inertial signals. Results are statistically significant ($p < 0.05$).}
    \centering
    \setlength{\tabcolsep}{4.5pt}
    \begin{tabular}{c|ccc|ccc}
        \hline
        \textbf{Method} & \textbf{Vision} & \textbf{Audio} & \textbf{IMU} & AUC@5 & AUC@10 & AUC@20 \\
        \hline
        Reloc3r \cite{reloc3r} & \checkmark &  &  & 0.57 & 2.33 & 7.33 \\
        Reloc3r \cite{reloc3r} w/ IMU & \checkmark &  & \checkmark & 0.79 & 3.00 & 8.83 \\
        Ours & \checkmark & \checkmark &  & 0.81 & 2.99 & 8.82 \\
        Ours w/ IMU & \checkmark & \checkmark & \checkmark & \textbf{0.85} & \textbf{3.04} & \textbf{8.90} \\
        \hline
    \end{tabular}
    \vspace{-8mm}
    \label{tab:imu}
\end{table}

\section{More Analyses}
\label{sec:additionalresults}

\subsection{Additional Model Analyses}
\label{sec:additionalmodelanalyses}
\noindent \textbf{Audio as a complementary signal to IMU for camera pose estimation.} In many practical perception systems, inertial sensors provide strong cues about camera motion. To better understand the role of spatial audio in such settings, we evaluate a variant of our model that incorporates IMU measurements alongside visual and audio inputs. Specifically, we follow the same late-fusion strategy used for audio features described in \cref{sec:pose_predictor} and incorporate IMU signals at the representation level. The IMU measurements are encoded and appended to the joint feature embedding formed by the visual and audio representations, resulting in a fused feature vector that combines visual, audio, and inertial information and is processed by the pose regression head. This design leaves the model architecture and training procedure otherwise unchanged, allowing us to directly compare vision-IMU and audio–visual–IMU variants.

As shown in \cref{tab:imu}, we see that adding IMU measurements improves the performance of our vision-only model, but combining our audio representation as an extra signal with the IMU enhances our model's performance even further. This indicates that passive scene sounds provide complementary spatial cues beyond visual and inertial signals. 
\medskip

\noindent \textbf{Impact of different audio clip lengths on model performance.} 
We experimented with different sound clip lengths to determine which ones work best for our method. The potential advantage of using a very short time window (e.g., 60 ms) is that the audio is dominated by a single instantaneous acoustic event, which can help the model more easily isolate a single sound source and avoid the mixing that happens when multiple events overlap. This also means there is less significant motion of sound sources during the capture window.

However, short clips are also limited, as they often contain too little acoustic variation to reveal meaningful spatial cues about the environment, especially in scenes where sound sources are quiet, intermittent, or dominated by background noise. In contrast, longer sound clips can capture more acoustic structure (reverberation, reflections, changes in relative loudness, and continuous sound dynamics) that provide stronger indirect signals about camera motion. These longer windows can help the model form a more stable representation of how the camera is moving relative to the surrounding scene.

We did not extend our experiments beyond a 1 second window for a few reasons. First, as clip length increases, the correspondence between the audio window and the ground-truth relative pose becomes increasingly ambiguous: within a long window, the camera may have moved significantly, making it unclear which portion of the audio should correspond to the target transformation. Second, longer windows increasingly entangle multiple independent sound events, which can blur the spatial cues that our model relies on. 

From our results in \cref{tab:auc20_clip_lengths}, we can see that using 1000ms clips consistently worked the best for all of our methods. 
\medskip

\begin{table}[t]
\centering
\caption{Total AUC@20 for different audio clip durations. Using 1000ms audio leads to the best performance for our setup.
}
\setlength{\tabcolsep}{4pt}
\begin{tabular}{l|ccc}
\toprule
\textbf{Sound Clip Size} & 60ms & 500ms & 1000ms \\
\midrule
\multicolumn{4}{l}{\textit{Audio Only}} \\
\midrule
NA-CP Model  & 0.68 & 0.69 & \textbf{0.72}\\
DOA CP Model & 0.68 & 0.72 & \textbf{0.78} \\
\midrule
\multicolumn{4}{l}{\textit{Audio + Vision}} \\
\midrule
Reloc3r \cite{reloc3r}-AV  & 7.83 & 7.38 & \textbf{8.07} \\
Ours w/o DOA   & 7.63 & 7.84 & \textbf{8.41} \\
Ours w/o Binaural &  7.48 & 8.01 & \textbf{8.34}  \\
Ours & 8.02 &  8.43 & \textbf{8.82} \\
\bottomrule
\end{tabular}
\label{tab:auc20_clip_lengths}
\end{table}

\begin{figure}
    \centering
    \includegraphics[width=0.8\linewidth]{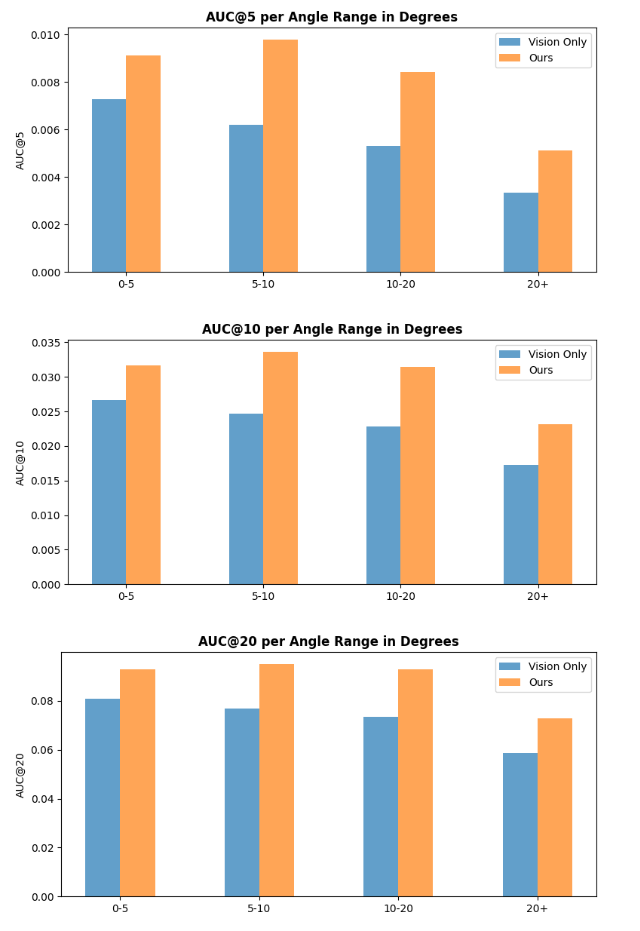}
    \caption{
    Our model performance compared against a SOTA vision-only baseline~\cite{reloc3r}. We evaluate our model's performance
    across different ranges of the magnitude of the ground-truth relative camera pose, expressed in angles, where these ranges are 0-5$^\circ$, 5-10$^\circ$, 10-20$^\circ$, and $>$20$^\circ$. We can clearly see that our model outperforms the baseline in all possible angle ranges.}
    \label{fig:angle-diffs}
\end{figure}

\noindent 
\textbf{Robustness to different magnitudes of target relative camera pose.}
Here, we evaluate our method against our vision-only model based on its ability to accurately predict estimates of target relative camera poses of different magnitudes. \Cref{fig:angle-diffs} shows the results from this analysis. We note that across all AUC thresholds, our method outperforms our SOTA vision-only baseline~\cite{reloc3r} for different ranges of target camera pose magnitude, expressed using angles,
including ranges with a mean greater than the overall mean magnitude of the ground-truth relative camera poses from our test set.
\medskip

\begin{table*}[t]
    \centering
    \tiny
    \caption{Total AUC scores across different temporal separations between frames in an input pair. Even though we perform our training uniformly using a 1 second difference between frames, our method generalizes across varying time differences. }
    \setlength{\tabcolsep}{4.5pt}
    \begin{tabular}{l|ccc|ccc|ccc|ccc}
        \hline
        \multirow{2}{*}{\textbf{Method}} 
            & \multicolumn{3}{c|}{\textbf{0.5s}} 
            & \multicolumn{3}{c|}{\textbf{1s}}
            & \multicolumn{3}{c|}{\textbf{2s}}
            & \multicolumn{3}{c}{\textbf{4s}}\\
        & @5 & @10 & @20 & @5 & @10 & @20 & @5 & @10 & @20  & @5 & @10 & @20\\
        \hline
        Reloc3r \cite{reloc3r} & 0.58 & 2.34 & 7.38 &  0.57 & 2.33 & 7.33  & 0.55  & 2.25  & 7.12 & 0.52 & 2.12  & 6.90  \\
        Ours  & \textbf{0.82} &\textbf{2.86} & \textbf{8.58 } & \textbf{0.81} & \textbf{2.99}& \textbf{8.82} & \textbf{0.70} & \textbf{2.78} & \textbf{8.74} & \textbf{0.66} & \textbf{2.64} & \textbf{8.63}\\
        \hline
    \end{tabular}
    \label{tab:temporal_splits}
\end{table*}

\noindent 
\textbf{Robustness to different inter-frame temporal separations.}
\Cref{tab:temporal_splits} compares our model's performance against Reloc3r \cite{reloc3r} across varying temporal gaps between the two frames in an input pair. Although our models are trained using frame pairs spaced one second apart, we find that this choice does not meaningfully limit generalization to larger temporal offsets. In evaluation, our audio-visual model not only outperforms the vision-only baseline across all AUC thresholds, but it also maintains stable performance as the frame gap increases, remaining consistent even at 4-second separations that were never seen during training. This indicates that our model can generalize to diverse frame rates in videos.
\medskip

\noindent \textbf{Behavior under strong visual conditions.} We determine reliability of vision signal by combining the use of Structural Similarity Index and ground truth relative camera pose changes. When vision is itself reliable (i.e., frames with minimal occlusion and large visual overlap), our model still outperforms the vision-only baseline in 95\% of such cases, with the remaining cases showing only a minimal performance gap. This is consistent with our late-fusion design (\cref{sec:pose_predictor}), which preserves the visual backbone's predictions and limits the extent to which uninformative audio can override an already-strong visual signal. These residual failures
occur predominantly when ground-truth camera motion is itself minimal,
where the absolute error gap between methods is negligible regardless of
which modality dominates. A smaller subset arises when audio changes
substantially despite negligible camera motion (e.g., a sound source
entering or growing louder). In these cases, the audio signal can pull the fused prediction away from the otherwise correct near-static vision-only estimate.

\begin{table}[t]
    \centering
    \caption{
    Mean of the ground-truth rotation and translation of the camera, expressed using angles, for different activity scenarios.
    }
    \setlength{\tabcolsep}{5pt}
    \begin{tabular}{l|c|c}
    \hline
        \textbf{Scenario} & \textbf{Mean Rotation ($^\circ$)}& 
         \textbf{Mean Translation ($^\circ$)} \\
         \hline 
         Basketball & 43.29 & 40.07 \\
         Bike Repair & 16.45 & 15.93 \\
         Cooking & 14.76 & 14.07 \\
         Dance & 40.98 & 40.38 \\ 
         Health & 8.12 &  9.24 \\
         Music & 7.29 & 8.09 \\
         Rock Climbing & 39.41 & 32.39 \\
         Soccer & 41.14 & 42.76 \\
         \hline
         Total & 18.35 & 17.94 \\
         \hline
    \end{tabular}
    \label{tab:rel-camera-pose-diffs}
    \centering
\end{table}

\newcommand{\heatcell}[2]{
  \cellcolor{#1!#2!white!50}
  \ifnum#2>50 \textbf{#2\%}\else #2\%\fi
}
\newcommand{\heatFF}[1]{\heatcell{red}{#1}}
\newcommand{\heatNF}[1]{\heatcell{blue}{#1}}
\newcommand{\heatDS}[1]{\heatcell{green}{#1}}
\newcommand{\heatLS}[1]{\heatcell{gray}{#1}}
\newcommand{\heatFC}[1]{\heatcell{orange}{#1}}
\begin{table*}[t!]
\centering
\small
\setlength{\tabcolsep}{4pt}
\begin{threeparttable}
\caption{Detailed distribution of settings and audio characteristics across
Ego-Exo4D scenarios (extends \cref{tab:ambient-audio}). \#Settings reports
the number of distinct recording sessions per scenario, each with
a different ambient sound template. FF/NF/DS/LS/FC report the percentage of frame
pairs exhibiting each audio characteristic (far-field sound source(s), near-field sound source(s), dominant single audio, low signal audio, and frequent sound Changes, respectively); column color matches the corresponding symbol in \cref{tab:ambient-audio}. Note: a frame pair may carry more than one tag. Cell shading indicates relative frequency (more saturated = more frequent, white = less frequent); bold values exceed 50\%, the threshold for a category marker in \cref{tab:ambient-audio}.}
\label{tab:audio-distribution}
\begin{tabular}{l c *{5}{>{\centering\arraybackslash}p{1.0cm}}}
\toprule
&  & \multicolumn{5}{c}{\% of Frame Pairs by Audio Characteristic\tnote{$\dagger$}}  \\
\cmidrule(lr){3-7}
Scenario & \#Settings & FF & NF & DS & LS & FC  \\
\midrule
Basketball    & 398 & \heatFF{84} & \heatNF{70} & \heatDS{10} & \heatLS{22} & \heatFC{60} \\
Bike Repair   & 249 & \heatFF{42} & \heatNF{74} & \heatDS{23} & \heatLS{61} & \heatFC{64} \\
Cooking       & 421 & \heatFF{38} & \heatNF{81} & \heatDS{35} & \heatLS{72} & \heatFC{17} \\
Dance         & 508 & \heatFF{41} & \heatNF{42} & \heatDS{82} & \heatLS{14} & \heatFC{59} \\
Health        & 287 & \heatFF{34} & \heatNF{65} & \heatDS{27} & \heatLS{78} & \heatFC{11} \\
Music         & 198 & \heatFF{11} & \heatNF{83} & \heatDS{83} & \heatLS{17} & \heatFC{42} \\
Rock Climbing & 133 & \heatFF{69} & \heatNF{56} & \heatDS{55} & \heatLS{25} & \heatFC{19} \\
Soccer        & 151 & \heatFF{74} & \heatNF{31} & \heatDS{32} & \heatLS{62} & \heatFC{55} \\
\bottomrule
\end{tabular}
\begin{tablenotes}
\tiny
\item[$\dagger$] Calculated from a random sample of 500 clip pairs manually annotated from each category, and extrapolated across the entire dataset.
\end{tablenotes}
\end{threeparttable}
\end{table*}

\subsection{Further Ego-Exo4D Dataset Analysis}
\label{sec:furtheregoexo}
As shown in \cref{tab:rel-camera-pose-diffs}, different scenarios exhibit widely varying amounts of camera motion. Activities like playing an instrument or performing COVID tests or CPR involve minimal movement, whereas sports and dynamic activities, such as basketball, soccer, rock climbing, and dancing, produce substantially higher motion. This highlights the broad diversity of scenarios captured in our dataset.

To annotate sound types for each scenario, we manually inspected subsets of videos (500 clips) within every scenario group and identified the dominant audio characteristics that occurred most frequently. Although individual videos may contain diverse and occasionally rare sound events, \cref{tab:ambient-audio} summarizes only the predominant acoustic characteristic observed within each scenario category. For example, in the Health scenario, we most commonly observed either minimal ambient audio or egocentric narration. In contrast, in scenarios like Basketball and Dance, we are often in much more reverberant environments such as gyms or dance studios, where we are more likely to either hear many different sound sources (from balls bouncing in the gym near or far away), or a dominant source that reverberates off the walls constantly (e.g. dance music playing in a dance studio). We applied this procedure consistently across all scenarios.

To quantify the dataset's underlying diversity, \cref{tab:audio-distribution} details the percentage of clips that exhibit specific audio characteristics within each scenario. It also reports the number of unique participant-environment pairs, accounting for the fact that different participants in the same location often experience distinct acoustic conditions at different times depending on what is happening. Notably, dynamic scenarios like Soccer and Bike Repair, both tagged as ``Frequent Changes'' in \cref{tab:ambient-audio}, rank among our highest-performing categories. This supports our findings in \cref{sec:results} that evolving audio content provides a more usable spatial signal for our method than a single static source.

\section{Implementation Details}
\label{sec:impdetails}
Here, we provide our model's training hyperparameters, based on the hyperparameters used to train the original Reloc3r \cite{reloc3r} network:
\begin{itemize}
    \item Training epochs: 100
    \item Warmup epochs: 5
    \item Learning rate: $10^{-5}$
    \item Minimum learning rate:  $10^{-7}$
    \item Batch size: 64
    \item Learning rate scheduler: Cosine annealing
\end{itemize}
\medskip
Here, we provide the steps and parameters for processing the visual inputs for Reloc3r~\cite{reloc3r}, specifically the Reloc3r-512 variant, which we use for all our experiments:
\begin{itemize}
    \item Image size: $256 \times 256$
    \item Data augmentation methods: Random color jittering
    \item Number of encoder layers: 24
    \item Number of decoder layers: 12
    \item Number of layers in pose regression head: 4
    \item Embedding size: 1024
\end{itemize}
\medskip
Here, we provide our audio processing parameters:
\begin{itemize}
    \item Audio sample rate: 48 kHz
    \item Spectrogram size per channel: $512 \times 96$
    \item Sound clip length: 1000ms 
    \item Number of FFT bins: 1024
    \item Number of sound sources we detect for DOA: 1
\end{itemize}
\medskip
Here, we provide details of our binauralization model (see \cref{sec:audio_enc} in main paper). We adopt the SLfM~\cite{slfm} architecture for the binauralizer,
but modify it by removing the visual encoder and using a ResNet-18 audio encoder with an embedding size of 1024 for the binaural audio features. The model also incorporates an audio U-Net \cite{vissound2.5}, following the design in SLfM \cite{slfm}.

To fuse our DOA spectrum and binaural embeddings, we first concatenate them into a single audio embedding of size $360+1024=1384$, where the feature size of the DoA and binaural embeddings are 360 and 1024, respectively. This joint audio embedding is then projected through a linear layer to produce a 768-dimensional audio representation. In parallel, the visual features extracted by the Reloc3r network are passed through their own linear layer to produce a 768-dimensional visual embedding. The transformed visual and audio embeddings are then concatenated to form a joint multimodal representation, which is then fed into Reloc3r’s pose regression head. Finally, this fused embedding is used to predict the camera pose matrix.

\section{Hardware and Resource Constraints}
\label{sec:hardware}

To train our model, we used 8 NVIDIA A40 GPUs, each with 48 GB of VRAM.

\section{Ethical Considerations}
\label{sec:ethics}
Because our method relies on incidentally recorded ambient audio, it raises important privacy considerations. In particular, using audio for camera-pose estimation could enable unintended forms of environmental or bystander tracking. Any real-world deployment of systems built on this technology should require explicit user consent and should ensure that individuals understand that both the visual and acoustic environments they encounter may be processed. Future work should incorporate safeguards such as on-device processing, strict data-retention limits, and transparent opt-in mechanisms to prevent misuse in surveillance-adjacent settings.


\end{document}